\documentclass[10pt,twocolumn,letterpaper]{article}

\usepackage[pagenumbers]{cvpr} 
\usepackage{float}
\usepackage{marvosym}
\usepackage{graphicx}
\usepackage{tabularx} 
\usepackage{amsmath}
\usepackage{amssymb}
\usepackage{booktabs}
\usepackage{tabulary,multirow,overpic,xcolor}
\usepackage{tikz,pgfplots}
\usepackage{siunitx}
\usepackage{tango}
\usepackage{ragged2e} 
\usepackage[pagebackref=true,breaklinks=true,letterpaper=true,colorlinks,
citecolor=citecolor2,bookmarks=false]{hyperref}
\usepackage[capitalize]{cleveref}
\usepackage{fontawesome5} 
\usepackage{pifont}
\newcommand{\specialcell}[2][c]{%
  \begin{tabular}[#1]{@{}c@{}}#2\end{tabular}}
\newcommand{\specialcellleft}[2][c]{%
	\begin{tabular}[#1]{@{}l@{}}#2\end{tabular}}

\newlength\savewidth
\newcommand\shline{\noalign{\global\savewidth\arrayrulewidth\global\arrayrulewidth 1pt}\hline\noalign{\global\arrayrulewidth\savewidth}}
\newcommand{\tablestyle}[2]{\setlength{\tabcolsep}{#1}\renewcommand{\arraystretch}{#2}\centering\footnotesize}

\definecolor{citecolor}{RGB}{34,139,34}
\definecolor{citecolor2}{HTML}{0071bc}
\definecolor{Graylight}{gray}{0.9}
\definecolor{lightred}{RGB}{241,140,142}

\begin{document}
	
	\title{  A Retrospective Systematic Study on Hierarchical Sparse Query Transformer-assisted Ultrasound Screening for Early Hepatocellular Carcinoma }
	\author{
	Chaoyin She\textsuperscript{ 1*} \qquad 
	Ruifang Lu\textsuperscript{ 2*} \qquad 
	Danni He\textsuperscript{ 2,3*} \qquad  
	Jiayi Lv\textsuperscript{ 4}\qquad
    Yadan Lin\textsuperscript{ 4} \qquad \\
	Meiqing Cheng \textsuperscript{ 2}\qquad
	Hui Huang\textsuperscript{ 2}\qquad
	Fengyu Ye\textsuperscript{ 5}\qquad
	Lida Chen\textsuperscript{ 2} \qquad
	Wei Wang\textsuperscript{ 2} \qquad
	Qinghua Huang\textsuperscript{ 1\Letter} \qquad \\ 
	\small   \\
	\textsuperscript{1}Northwestern Polytechnical University 
	\\
	 \textsuperscript{2}The First Affiliated Hospital of Sun Yat-Sen University
	 \\
	 \textsuperscript{3}The Seventh Affiliated Hospital of Sun Yat-Sen University
	 \\
	 \textsuperscript{4}The First Affiliated Hospital of Guangxi Medical University
	 \\
	 \textsuperscript{5}Xi'an Jiaotong University
}
	\maketitle
	
\begin{abstract}
Hepatocellular carcinoma (HCC), ranking as the third leading cause of cancer-related mortality worldwide, demands urgent improvements in early detection to enhance patient survival. While ultrasound remains the preferred screening modality due to its cost-effectiveness and real-time capabilities, its sensitivity (59\%-78\%) heavily relies on radiologists' expertise, leading to inconsistent diagnostic outcomes and operational inefficiencies. Recent advancements in AI technology offer promising solutions to bridge this gap. This study introduces the Hierarchical Sparse Query Transformer (HSQformer), a novel hybrid architecture that synergizes CNNs’ local feature extraction with Vision Transformers’ global contextual awareness through latent space representation and sparse learning. By dynamically activating task-specific experts via a Mixture-of-Experts (MoE) framework, HSQformer achieves hierarchical feature integration without structural redundancy. Evaluated across three clinical scenarios—single-center, multi-center, and high-risk patient cohorts—HSQformer outperforms state-of-the-art models (e.g., 95.38\% AUC in multi-center testing) and matches senior radiologists’ diagnostic accuracy while significantly surpassing junior counterparts. These results highlight the potential of AI-assisted tools to standardize HCC screening, reduce dependency on human expertise, and improve early diagnosis rates. The full code is available at \url{https://github.com/Asunatan/HSQformer}.
	
\end{abstract}

\renewcommand*{\thefootnote}{\fnsymbol{footnote}}
\footnotetext{* Equal contribution.}
\renewcommand{\thefootnote}{\fnsymbol{footnote}}
\footnotetext{\Letter  Corresponding Author.}

	\section{Introduction}
	\label{sec:intro}
Liver cancer ranks as the sixth most common malignancy and the third leading cause of cancer-related mortality globally, with hepatocellular carcinoma (HCC) accounting for approximately 75\% - 85\% of primary liver cancers\cite{globalcancer}. Early detection and treatment of HCC can improve patient survival rates and life expectancy. Ultrasound (B-mode) is the preferred imaging modality for screening high-risk populations for HCC. Compared to CT or MRI, ultrasound offers the advantages of portability, radiation-free imaging, and real-time visualization. However, according to meta-analyses, the sensitivity of conventional B-mode ultrasound in detecting HCC ranges from only 59\% to 78\%, which is highly dependent on the clinical experience of radiologists\cite{metaanalysis1,metaanalysis2}. Moreover, ultrasound examinations require substantial time from these specialists for image interpretation and secondary confirmation, often making clinical assessments inefficient. These limitations underscore the urgent need for robust tools to standardize HCC screening and reduce reliance on subjective human interpretation.

Recent advancements in artificial intelligence (AI), particularly deep learning-based computer-aided diagnosis (CAD) systems, demonstrate transformative potential for improving clinical decision-making in medical image interpretation. 
	\begin{figure}[t]
		\centering
  \vspace{-10pt}
		\includegraphics[width = \linewidth]{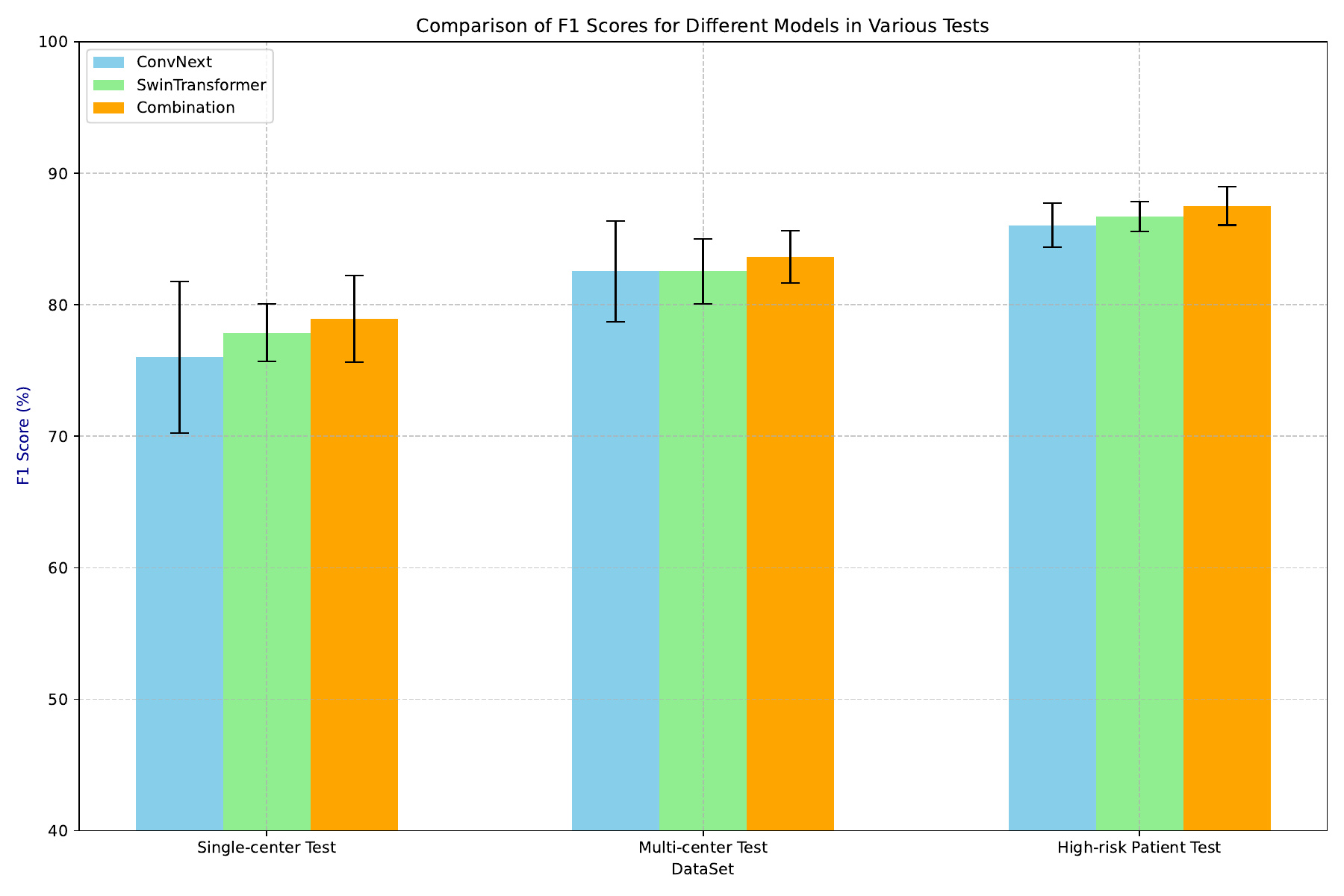}
\caption{The improvement brought by simply combining CNN and ViT.
}
		\label{fig1}
  \vspace{-10pt}
  	\end{figure}
Current AI-based CAD frameworks predominantly leverage convolutional neural networks (CNNs) or vision transformers (ViTs)\cite{vit}. CNNs have been pivotal in computer vision tasks due to their ability to capture local texture information, but their performance is often limited by the size of their receptive fields, restricting their ability to perceive long-range dependencies and integrate global contextual information. ViTs, on the other hand, excel in capturing global semantic information through self-attention mechanisms but typically require extensive datasets and significant computational resources for effective pre-training, and lack the inductive biases inherent in CNNs.

 This raises the question: is there a straightforward method to address the limitations of these two architectures? A simple yet feasible solution is to combine the strengths of CNNs and ViTs to form a hybrid model. However, current integration methods often necessitate complex architectural modifications or the use of sophisticated design techniques. Is it possible to achieve effective complementarity between CNNs and ViTs without altering the network structure? This task presents significant challenges. As shown in Figure \ref{fig1}, the two most popular network architectures, ConvNext (CNNs)\cite{ConvNext} and SwinTransformer (ViTs)\cite{swin}, were selected for preliminary experiments. Our results show that late fusion\cite{Hybridsurvey} of CNNs and ViTs offers limited benefits, as this approach tends to overemphasize high-level semantics for classification while neglecting low-level details and textures. Moreover, this approach\cite{Hybridsurvey} fails to leverage multi-scale features, which have been proven crucial for downstream tasks such as segmentation and detection. 
 
To address these limitations and bridge the gap between the two architectures, we propose the Hierarchical Sparse Query Transformer (HSQformer), a novel hybrid architecture that hierarchically synergizes the local feature extraction capabilities of CNNs with the global contextual awareness of ViTs through latent space representation and sparse learning. The HSQformer dynamically activates task-specific experts via a Mixture-of-Experts (MoE) framework, achieving hierarchical feature integration without structural redundancy. Evaluated across three clinical scenarios—single-center, multi-center, and high-risk patient cohorts—the HSQformer demonstrates superior performance compared to state-of-the-art models and matches the diagnostic accuracy of senior radiologists while significantly surpassing that of junior counterparts. This study represents the first systematic exploration of AI-assisted ultrasound screening for HCC, offering a robust and standardized diagnostic tool to improve early detection rates and patient outcomes.
%

In summary, our contributions are as follows:
\begin{itemize}
	\item We introduce the novel Hierarchical Sparse Query Transformer, which integrates the strengths of CNNs and ViTs without necessitating complex modifications, adhering to a modular and extensible design philosophy.
	\item To our best knowledge, our work represents the first large - scale validation of AI - assisted ultrasound screening for HCC across distinct clinical environments, ensuring generalizability and practical relevance.
	\item We performed a human-machine comparative analysis to validate the potential application of artificial intelligence tools in ultrasound-assisted early HCC screening.
	\item We have open-sourced our code and checkpoints, and established the pioneering benchmark for AI-assisted ultrasound screening of HCC, which is accessible at \url{https://paperswithcode.com/sota/classification-on-liver-us}. 
\end{itemize}
\section{Related Work}
\label{sec:related}
\subsection{Convolutional Neural Networks}
Convolutional Neural Networks (CNNs) have revolutionized computer vision, serving as a cornerstone for tasks such as image classification, object detection, and semantic segmentation \cite{resnet,densenet}. Their success stems from their ability to autonomously learn hierarchical features directly from raw pixel data, eliminating the need for manual feature engineering. Key inductive biases, including spatial locality and translational invariance, enable CNNs to efficiently capture local patterns while maintaining robustness to positional variations in natural images.

Despite these strengths, conventional CNNs exhibit limitations in medical imaging applications. Their restricted receptive fields hinder the integration of global contextual information, which is critical for interpreting complex anatomical structures and subtle pathological features [3]. This constraint becomes particularly detrimental when local regions require semantic dependencies spanning the entire image.Recent advancements aim to enhance CNN generalizability through three synergistic strategies: (1) Expanding receptive fields via enlarged kernels \cite{RepLKNet,slak,pelk} or structural re-parameterization\cite{acnet,repvgg,repvit}; (2) Incorporating adaptive operators such as dilated\cite{dilated} or deformable convolutions \cite{deformablev1,deformablev2,internimage} to dynamically adjust spatial sampling; (3) Integrating attention mechanisms\cite{senet,cbam} to refine feature discriminability. These innovations preserve CNNs' local feature extraction strengths while enhancing global contextual modeling, positioning them as adaptable frameworks for medical image analysis

\subsection{Vision Transformers}
In natural language processing (NLP), Transformers\cite{transformer} have demonstrated exceptional capabilities in modeling long-range dependencies. This success has inspired researchers to explore the application of Transformer architectures to computer vision, where the ViT represents a landmark development. Recent works have explored applying ViT to various vision tasks: image classification, object detection, image segmentation, depth estimation, image generation, video processing, and others. Despite its success, ViT also presents certain limitations. One notable drawback is its reliance on large-scale datasets for effective training, as Transformers lack the inductive biases inherent in CNNs, such as locality and translation invariance\cite{cvt}. As a result, ViTs may struggle to generalize effectively when trained on smaller datasets\cite{cct}. Additionally, ViTs are computationally demanding, particularly due to their self-attention mechanisms, which scale quadratically with image size, leading to high memory and processing costs. 

To enhance the efficiency of ViTs while maintaining performance, several studies\cite{swin,slide,cswin,p2t,pvtv2,metaformer} have proposed methods that incorporate local self-attention mechanisms or pooling operations. By restricting self-attention to localized regions of an image or employing pooling to condense feature maps, these optimizations not only streamline processing but also ensure that the models remain adept at handling diverse vision tasks, thereby achieving a balance between computational efficiency and model efficacy. Several other studies have primarily concentrated on diversifying the ViT architectures and enhancing their adaptability, such as the Focal Transformer\cite{focal}, MobileViT\cite{mobilevit}, and Axial-Attention Transformer\cite{valanarasu2021medical}, each contributing to the field with innovative approaches to attention mechanisms and knowledge distillation techniques.
\subsection{Latent Space and Sparse Learning}
The concept of Latent Space is pivotal in machine learning for capturing the intrinsic features of data. Its utility in deep learning models is exemplified by intermediate layer representations, which are crucial for various tasks\cite{goyal2021coordination}. To enhance efficiency and prediction accuracy, learned queries have been explored in architectures such as SetTransformers\cite{Settransformer} and Perceiver \cite{Perceiver} networks, where they project inputs into a lower-dimensional space. Goyal et al.\cite{goyal2021coordination} further advanced this concept by leveraging learned queries as a shared workspace to reduce computational complexity in Transformers. In addition to these developments, attention mechanisms have also been optimized through the incorporation of learnable tokens. Models such as Involution\cite{Involution}, VOLO\cite{Volo}, and QnA\cite{Learnedqueries} demonstrate how learnable tokens can replace traditional queries or keys, generating dynamic affinity matrices that significantly boost model performance. Collectively, these innovations underscore the importance of Latent Space in driving the evolution and enhancing the capabilities of deep learning models.

Sparse learning stands as a pivotal approach within machine learning, particularly for its effectiveness in optimizing computational resources and enhancing model performance. A prime example of this approach is the Mixture-of-Experts (MOE) framework, which partitions models into specialized experts, selectively activating only the relevant subset to process inputs and thereby achieving computational efficiency. The Multi-Mixture-of-Experts (MMoE)\cite{mmoe} model epitomizes MOE's application in multi-task learning, designed to explicitly learn task relationships from data. It enables the sharing of expert sub-models across various tasks while simultaneously training a gating network to optimize performance for each individual task. Within the realm of large language model(LLM), MOE has attracted considerable scholarly attention; studies such as DeepSeek\cite{deepseekv1}, and LLaMA-MoE\cite{LlamamoeV1,LlamamoeV2} are at the forefront of research, showcasing the potential of MOE in scaling up large language models  with a consistent number of activated parameters. In summary, MOE emerges as a potent sparse representation model, and it is recommended that readers consult reference \cite{moesurvey} for an in-depth exploration of the subject.
	\section{Approach}
	\label{sec:method}
\begin{figure*}[t]
	\vspace{-15pt}
	\centering
	\includegraphics[width = \linewidth]{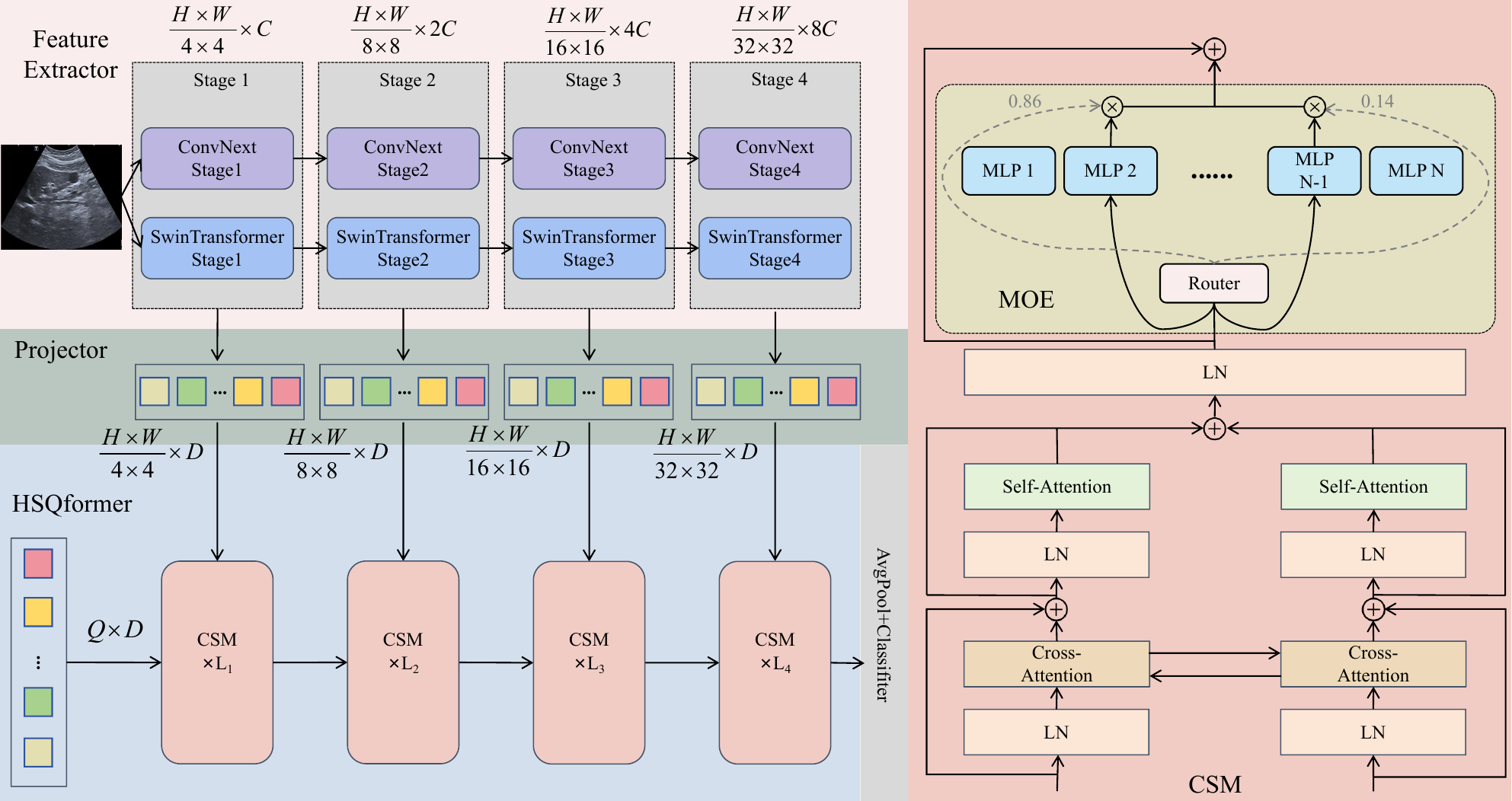}
	\caption{\textbf{model architecture}. Overview of the HSQformer, integrating CNN and ViT features through a hierarchical sparse querying framework for efficient diagnosis.
	}
	\label{fig2}
	\vspace{-10pt}
\end{figure*}
	\subsection{Overall Architecture}
Our objective is to design an architecture that effectively combines the complementary strengths of CNNs and ViTs, without any unnecessary bells-and-whistles design. This architectural framework is deployed in real-world clinical ultrasound screening for hepatocellular carcinoma (HCC) to investigate the potential of AI-assisted diagnosis in a clinical context. To achieve this goal, we introduce a paradigm that leverages latent space representation and sparse learning to integrate these strengths. As illustrated in Figure \ref{fig2}, our architecture consists primarily of a feature extractor, a projector, and a HSQformer. 
The HSQformer is a novel multi-level sparse Q-former, composed of stacked hierarchical learnable Query Transformers. We selected the state-of-the-art ConvNeXt and SwinTransformer as feature extractors due to their simplicity and efficiency. Similar to ConvNeXt and SwinTransformer, the HSQformer is organized into four stages that aggregate feature maps at various scales. Each stage shares a similar modular design, consisting of multiple Cross-Self-attention Mixed experts (CSM) blocks, which enables scalability and reusability.
	
Given an input image $X\in \mathbb{R}^{C \times H \times W}$, we first apply two distinct feature extractors to capture multi-scale representations. The resulting feature maps are denoted as $\left \{ f_{c1},f_{c2},f_{c3},f_{c4}\right \}$ for ConvNeXt and $\left \{ f_{s1},f_{s2},f_{s3},f_{s4}\right \}$ for SwinTransformer. Each set of feature maps is generated at different spatial resolutions with strides ($S$) of 4, 8, 16, and 32 pixels, respectively, relative to the original image pixels.

The multi-scale feature maps, along with a level embedding, are simultaneously fed into a projector module comprising four CSM modules. The projector module is designed to project the features uniformly into a $D$-dimensional latent space. This process reduces redundancy within the feature representations and produces a set of compact and efficient latent space representations, which are essential for effectively capturing the critical characteristics of the input data.

The latent space features are hierarchically injected into the HSQformer to extract salient features related to diagnosis. Subsequently, a linear projection layer is applied for the classification. This approach adheres to the standard backbone network design, ensuring the rationality and feasibility of the network architecture.

\subsection{HSQformer}
\label{sec:method:HSQformer}
Inspired by references \cite{Learnedqueries,mae,mmoe}, we introduce latent space representation and sparse learning to construct a novel Hierarchical Sparse Querying Transformer, termed HSQformer. This architecture comprises a learnable query embedding representation and a four-stage backbone, with each stage composed of Cross-Self-attention Mixed experts (CSM) modules, facilitating the scalability of the HSQformer. The size of the HSQformer can be varied by adjusting the number of CSM modules in each stage, resulting in different configurations such as Small, Base, and Large with detailed specifications provided in Appendix.

There are two major differences between the HSQformer and the Q-former: 
\begin{itemize}
	\item \textbf{Sparse Optimization}. In contrast to Q-former, which processes input tokens densely, HSQformer utilizes tokens from a latent space and integrates MoE for sparse processing. This enables HSQformer to activate only the most pertinent experts, thereby improving computational efficiency and scalability.
	\item \textbf{Hierarchical Querying}. the Q-former's querying is confined to the final layer's high-level semantics, but the HSQformer adopts a hierarchical querying strategy that encompasses both low-level and high-level features. 
\end{itemize}
\subsubsection{CSM}
\label{sec:method:CSM}
As illustrated in Figure \ref{fig2}, the Cross-Self-attention Mixed experts (CSM) consists of cross-attention, self-attention, and a mixture of experts, closely aligning with the standard transformer architecture without incorporating unnecessary bells-and-whistles. The role of CSM varies depending on its specific location within the model. When situated within the projector, the CSM facilitates the integration of fine-grained local features with global contextual information, and projects the resulting representations into a $D$-dimensional latent space for subsequent processing. In contrast, when integrated into the HSQformer, the CSM primarily leverages query embeddings to  extract features from the latent space that are highly pertinent to diagnostic classification tasks, ensuring that the most relevant information is captured for accurate decision-making.

Specifically, Cross-Attention (CA) within the CSM facilitates the interaction between inter-feature representations, enabling the model to capture complex interdependencies. Self-Attention (SA) is subsequently applied to refine intra-feature representations, further enhancing the internal coherence of features post-interaction. The Mixture of Experts (MOE) introduces sparse representations by dynamically routing each token to a subset of specialized experts, effectively increasing the model's capacity while preserving computational efficiency through selective activation. 

Note that the CSM structure is symmetric; for simplicity, only one side is discussed here, with the other half being analogous.
Assuming that $x_{scr}$  and $x_{tgt}$ are the feature representations of the source sequence and the target sequence respectively, the basic principle can be roughly described as follows:	
\begin{equation}
	\label{eq:CSM}
	C S M\left(x_{s c r}, x_{t g t}\right) = MOE\left(SA \left(C A\left(x_{s c r}, x_{t g t}\right)\right) \right)\\
\end{equation}
\begin{equation}
	\label{eq:SA}
	S A\left(x_{s c r}\right)  = Softmax(\frac{x_{s c r} W_{Q}\left(x_{s c r} W_{K}\right)^{T} }{\sqrt{d} } )x_{s c r}W_{V} 
\end{equation}
\begin{equation}
	\label{eq:CA}
	C A\left(x_{s c r}, x_{t g t}\right)  =  Softmax(\frac{x_{s c r} W_{Q}\left(x_{t g t} W_{K}\right)^{T} }{\sqrt{d} } )x_{t g t}W_{V} 
\end{equation}
where $W_{Q}$, $W_{K}$, and $W_{V}$represent the weight matrices for Query, Key, and Value, respectively; $d$ denotes the dimensionality of the query and key features, which is used to scale the dot product in order to mitigate the vanishing gradient problem. 

\subsubsection{Latent Space Representation and MOE}
\textbf{Latent Space Representation}.
Considering the feature map $f_{l_{i} } \in \mathbb{R}^{N_{i} \times D}(i\in\left [  1,2,3,4\right ]  )$ in the latent space, where $N_{i}$, the sequence length of stage $i$, is calculated as $\frac{H\times W}{S_{i}^{2} }$, and $D$ represents the embedding dimension. Corresponding query embeddings can be denoted as $f_{q } \in \mathbb{R}^{Q \times D}$, where $Q$ denotes the number of query tokens. The interaction between query tokens and latent space features can be formulated as:
\begin{equation}
	\mathcal{O}\in \mathbb{R}^{Q\times D}   = \mathcal{F}(f_{q}\in \mathbb{R}^{Q \times D},f_{l_{i}}\in \mathbb{R}^{N_{i} \times D}  )
\end{equation}
where the function $\mathcal{F}$ encapsulates the computational process that is analogous to the expressions embodied in Equations \ref{eq:CSM}, \ref{eq:SA} and \ref{eq:CA}. Notably, the sequence lengths $N_{i}$ for the first two stages are relatively large (e.g., 3136), indicating that these feature maps contain more detailed and low-level features, which may include redundant information. Conversely, the latter two stages exhibit smaller lengths (e.g., 49), typically reflecting abstract high-level semantics. By judiciously selecting $Q$, we can effectively reduce feature redundancy and enhance semantic richness, thereby optimizing the model's ability to identify features that are crucial for diagnostic classification. This enhancement is empirically validated through a series of comprehensive ablation studies.

\textbf{MOE}. The $Q$ tokens are processed through the MOE module, which selectively activates the \textit{top-k} experts using a routing mechanism. This approach dynamically allocates input tokens to different experts, allowing each to focus on their specific area of expertise, thereby enhancing the model's efficiency and promoting sparsity by engaging only a subset of experts during inference. In this study, each expert network $F_{i}(i \in [1,E])$ is implemented as a Multilayer Perceptron (MLP) network, which accepts an input $x$ and produces an output $F_{i}(x)$. Concurrently, a gating network $G$, which is composed of an MLP followed by a softmax layer, generates the output $G(x)$. The output of the MoE
layer can be formulated as
\begin{equation}
	MOE(x)=\sum_{i=1}^{E} G(x)_{i} F_{i}(x)
\end{equation}
\begin{equation}
	G(x)_{i}=\frac{\exp(TopK(xW+R_\mathrm {noise})_{i})}{\sum_{j=1}^{E} \exp (TopK(xW+R_\mathrm {noise})_{j})}
\end{equation}

\begin{equation}
	TopK(xW + R_{\text{noise}})_i = \left\{
	\begin{array}{ll}
		(xW + R_{\text{noise}})_{i}, & \text{if } condition, \\
		-\infty, & otherwise.
	\end{array}
	\right.
\end{equation}
The essence of MOE is a form of conditional computation. In this context, the term $condition$ denotes that $(xW + R_{\text{noise}})_{i}$ is ranked among the \textit{top-k} elements within the ensemble of $xW + R_{\text{noise}}$. Here, $W$ represents the weight matrix of the gating network.

\section{Experiments}
\subsection{Experimental Setup}
\paragraph{Datasets.}The dataset in this study was sourced from the multi-center dataset of the Ultrasound Engineering Society of China's Medical Industry Branch (UE-MICAP). It contains 11,149 cases with 19,464 images collected from January 2014 to December 2023. The data were obtained from these hospitals: the First, Third, Sixth, and Seventh Affiliated Hospitals of Sun Yat-sen University; the First Affiliated Hospital of Guangzhou Medical University; the First Affiliated Hospital of Guangxi Medical University; Foshan Sanshui District People's Hospital; and West China Xiamen Hospital of Sichuan University.

The inclusion criteria for the study were: (a) patients at risk of HCC, including those with clinical diagnoses or imaging evidence of cirrhosis, or pathological confirmation of cirrhosis; (b) males over the age of 40 and females over the age of 50 with a history of viral hepatitis; (c) patients who underwent ultrasound screening and had lesions confirmed by clinical or pathological diagnosis. The exclusion criteria were: (a) individuals under the age of 18; (b) images with inadequate quality; (c) images containing both benign and malignant lesions, as this could potentially impair the AI's ability to accurately classify the lesions. This retrospective study utilized fully anonymized images, thus eliminating the need for informed consent.

Table \ref{tab1} presents the three testing scenarios: single-center test, multi-center test, and high-risk patient test. The single-center test is an independent evaluation using a dataset separate from the training set. For the multi-center test, data from eight hospitals were combined to mimic a diverse clinical setting. The high-risk patient test focuses on patients with hepatitis, assessing model performance in this challenging subgroup. These scenarios mirror real-world clinical applications, offering a comprehensive evaluation of the model's effectiveness and robustness across different conditions.
\begin{table}[t!]
	\centering
	\tablestyle{1.8pt}{1.05}
	\resizebox{1.04\linewidth}{!}{
		\begin{tabular}{l|c|c|c|c}
			\multicolumn{1}{c|}{parameters}  &   Training  & \specialcell{ Single center\\ Test}  & \specialcell{Multi center \\ Test} & High risk \\
			\shline
			Number of patients  & 6376 & 635 & 2684 & 1454    \\
			Number of images       & 11380 & 635 & 4565 & 2884    \\
			Benign  images   &  8225 & 200 & 2295 & 706   \\
			Malignant images   &  3155  & 435  & 2270 & 2178   \\ 
			Data sources        &  \specialcell{ the First Affiliated\\ Hospital of Sun \\Yat-sen University}   & \specialcell{ the Third Affiliated\\ Hospitalof Sun  \\Yat-sen University}  & UE-MICAP & \specialcell{ the First Affiliated\\  Hospital of Sun\\ Yat-sen University}    \\ 
		\end{tabular}
	}
	\caption{The chart depicts the distribution of benign and malignant cases, with data sourced from multiple medical institutions. 
	}
	\label{tab1}
\end{table}
\begin{table*}[h]
	\begin{center}
		\tablestyle{1.8pt}{1.05}
		\resizebox{1.0\linewidth}{!}{
			\begin{tabular}{l|lcccr|lcccr|lcccr}
				
				\multirow{2}{*}{Models} & \multicolumn{5}{c|}{Single-Center Test} & \multicolumn{5}{c|}{Multi-Center Test} & \multicolumn{5}{c}{High-Risk Patient Test}  \\
				\cline{2-16}
				~ &Accuracy  & Precision & Recall&F1 & \multicolumn{1}{c|}{AUC}
				& Accuracy & Precision & Recall&F1  & \multicolumn{1}{c|}{AUC}
				& Accuracy  & Precision& Recall&F1  & \multicolumn{1}{c}{AUC}
				\\
				\shline
				ResNet101\cite{resnet} 
				&72.38±0.63&	85.17±3.43&	72.6±4.54&78.22±1.25 &	79.7±2.25 
				&83.52±0.78&	89.1±1.78&	76.81±2.73&	82.46±1	&	93.02±0.53
				&81.6±1.77&	89.17±1.7&	86.21±4.77&	87.57±1.61&	86.69±0.43\\
				DenseNet201\cite{densenet} 
				&71.34±1.27&82.54±5.14&	74.76±8.61&	77.98±2.49&	78.51±2.43
				&84.22±2.27&	87.84±2.28&	80.07±7.44&	83.54±3.24&	93.3±0.51
				&81.2±2.23&	88.86±2.22&	86.02±5.08&	87.31±1.9&	85.83±1.78\\
				ResNext101\cite{resnext} 
				&74.74±3.69&	83.9±1.68&	78.25±7.63&	80.78±3.83&	80.34±2.69
				&84.39±1.44&	88.5±1.64&	79.48±4.26&	83.67±1.92&	93.36±0.49
				&81.15±0.91&	88.65±1.7&	86.12±1.52&	87.34±0.55&	86.26±2.19
				\\
				ThyNet\cite{ThyNet} 
				&72.21±2.16&	83.34±3.76&	74.99±9.06&	78.52±3.26&	79.61±0.68
				&83.36±1.91&	88.23±3.22&	77.79±7.96&	82.37±2.95&	93.25±0.63
				&80.69±3.45&	89.51±1.54&	84.44±7&	86.73±2.96&	86.54±0.99
				\\
				Hiera\cite{hiera} 
				&72.44±3.13&	83.41±1.88&	74.57±3.47&	78.73±2.66&	78.51±2.67
				&84.79±0.84&	89.37±1.11&	79.32±2.04&	84.03±1.04&	93.8±0.53
				&80.84±0.58&	88.92±0.72&	85.26±1.57&	87.04±0.53& 85.47±0.76
				\\
				ViT\cite{vit} 
				&77.23±1.94&	80.1±1.66 &	88.97±4.66&	84.22±1.75&	81.37±1.31
				&85.55±0.62&	86.16±1.39&	85.06±2.76&	85.57±0.94&	93.57±0.45
				&84.02±1.35&	87.22±0.88&	92.4±2.19 &	89.72±0.97&	86.44±0.83
				\\
				SwinTransformer\cite{swin} 
				&74.74±2.69&	85.2±1.81 &	76.6±7.05 &	80.48±2.93&	82±0.28
				&85.39±1.08&	89.59±1.65&	80.45±3.43&	84.72±1.5 &	94.15±0.63
				&81.66±1.19&	89.09±0.83&	86.31±2.25&	87.66±0.95&	86.49±1.05
				\\
				CswinTransformer\cite{cswin} 
				&74.4±2.09&	\textcolor{red}{85.77±2.32}&	75.31±6.07&	80.01±2.71&	82.35±0.71
				&85.22±1  &	90.07±1.54&	79.54±2.94&	84.43±1.29&	94.33±0.46
				&80.63±1.94&	\textcolor{red}{90.38±1.51}&	83.29±4.13&	86.62±1.77&	86.86±0.87
				\\
				ConvNext\cite{ConvNext} 
				&76.44±1.67&	84.85±3.22&	80.27±6.78&	82.27±2.02&	83.41±1.07
				&86.39±1.11&	89.18±2.78&	83.38±5.53&	86.02±1.6&	94.45±0.53
				&82.96±1.65&	88.64±1.66&	88.93±4.55&	88.7±1.46&	86.94±0.99
				\\
				PVTv2-B5\cite{pvtv2} 
				&76.47±1.2&	83.93±2.37&	81.47±5.54&	82.53±1.62&	83.1±1.52
				&85.9±0.75&	\textcolor{red}{90.08±2.51}&	81.15±4.08&	85.28±1.16&	94.44±0.37
				&81.88±2.04&	89.56±1.4&	86.12±4.34&	87.73±1.69&	87.02±0.94
				\\
				FocalNet\cite{focalnet} 
				&76.98±1.99&	83.71±1.56&	82.57±5.53&	83.02±2.11&	82.16±1.98
				&85.85±1.55&	89.78±2&	81.36±5.4&	85.24±2.11&	94.45±0.32
				&82.57±1.48&	89.27±1.92&	87.58±4.59&	88.32±1.37&	87.07±0.52
				\\
				MPViT\cite{mpvit} 
				&75.94±2.59&	83.8±2.86&	80.78±7.4&	82.02±2.86&	81.9±1.78
				&84.82±1.95&	89.75±2.16&	79.15±5.97&	83.96±2.58	&93.97±0.49
				&81.03±2.41&	89.55±1.59&	84.85±4.38&	87.07±1.97&	85.5±2.27
				\\
				\hline
				HSQformer-S
				&76.53±2.4&	82.02±2.47&	84.55±7.58&	83.04±2.71&	82.07±1.66
				&87.41±0.55&	85.56±1.95&	90.44±3.49&	87.87±0.76&	94.71±0.39
				&84.44±0.33&	86.5±1.27&	94.12±2.17&90.13±0.35&	86.64±0.64
				\\
				HSQformer-B 
				&78.74±3.05&	80.29±4.57&	\textcolor{red}{92.14±4.71}&	85.62±1.47&	\textcolor{red}{83.83±0.96}
				&88.29±0.71&	87.09±2.7&	90.38±4.13&	88.6±0.85&	\textcolor{red}{95.38±0.33}
				&\textcolor{red}{85.41±0.58}&	87.62±2.07&	94.09±2.75&	90.69±0.38&	\textcolor{red}{88.32±0.59}
				\\
				HSQformer-L
				&\textcolor{red}{79.62±0.98}&	81.06±1.53&	91.81±4.14&	\textcolor{red}{86.03±1.12}&	83.41±1.39
				&\textcolor{red}{88.51±0.49}&	86.51±1.62&	\textcolor{red}{91.57±2.37}&	\textcolor{red}{88.94±0.55}&	95.04±0.51
				&85.38±0.6&	87.05±1.36&	\textcolor{red}{94.79±2.18}&	\textcolor{red}{90.73±0.47}&	88.23±1.08
				\\
			\end{tabular}
		}
		\caption{\textbf{SOTA Model Comparison}. This table presents a comprehensive comparison of our proposed model with existing state-of-the-art CNNs and ViT models, highlighting their respective performances and characteristics.
		}
		\label{tab2}
	\end{center}
\end{table*}
\paragraph{Implementation details.} 
The original DICOM images were losslessly converted to PNG format via OpenCV for standardization. Then, textual metadata like patient identifiers, device information, institutional details, and date stamps were removed using the YOLO object detection model\cite{yolov3}. We followed the fine-tuning protocol from CSWinTransformer\cite{cswin} for fair comparison. A 5-fold patient-level cross-validation strategy was implemented to prevent data leakage and ensure result robustness. Model weights were saved after each epoch with AUC - based validation performance improvement, without early stopping to avoid underfitting or overfitting. More details are in the appendix.
\subsection{Results}
\subsubsection{SOTA Model Comparison}
\paragraph{Single-Center Test.}
In the single-center Test scenario, the HSQformer-B demonstrates superior Recall (92.14±4.71 \%) and AUC (83.83±0.96 \%) compared to other models, indicating its exceptional ability to correctly identify positive cases within a single institutional dataset. The HSQformer-S also shows competitive performance, with an AUC of 82.07±1.66\%, surpassing models such as ResNet101\cite{resnet}, DenseNet201\cite{densenet}, and ThyNet\cite{ThyNet}.
\paragraph{Multi-Center Test.}
Transitioning to the multi-center test, which evaluates model generalizability across different institutions, HSQformer-B achieves leading scores in Accuracy (88.29±0.71\%), AUC (95.38±0.33\%), and Recall (90.38±4.13\%). This suggests that HSQformer not only excels in identifying positive cases but also maintains high consistency across diverse datasets. ConvNext has strong overall performance, particularly in Accuracy (86.39 ± 1.11\%) and AUC (94.45 ± 0.53\%), but falls short when compared to HSQformer.
\paragraph{High-Risk Patient Test.}
In the high-risk patient test, which is designed to evaluate the model's effectiveness on patients identified as high-risk, HSQformer-B once again achieves top-tier metrics: Accuracy (85.41±0.58\%), AUC (88.32±0.59\%), and F1 score (90.69±0.38\%). Notably, the Recall of 94.09±2.75\% underscores its proficiency in detecting critical cases. ViT, while performing well with an Accuracy of 84.02±1.35\% and Recall of 92.4±2.19\%, does not match the comprehensive superiority of HSQformer-B.
\subsubsection{Radiologists vs. AI} 
\begin{table}[h]
	\centering
	\tablestyle{1.8pt}{1.05}
	\resizebox{1.04\linewidth}{!}{
		\begin{tabular}{l|c|c|c|c|c}
			&  Accuracy  & Precision & Recall
			&F1 & \multicolumn{1}{c}{AUC} \\
			\shline
			senior 1 & 75.04 & 94.72 & 70.89 & 81.09 & 79.4\\
			senior 2 & 82.77 & 90.94 & 85.72 & 88.25 & 79.7\\
			senior 3 & 80.41 & 89.63  & 83.75& 86.59 & 76.9\\ 
			senior average & 79.41±3.96&\textcolor{red}{91.76±2.64}&	80.12±8.05&85.31±3.75&78.67±1.54\\ 
			\hline		
			junior 1 & 79.61& 86.64 & 86.32 &86.48& 72.6 \\ 
			junior 2 & 77.22 &82.97 & 87.88 &85.35 & 66.1 \\ 
			junior 3 & 78.66 & 86.8& 84.81 &85.79 &  72 \\ 
			junior average  & 78.5±1.2&	85.47±2.17&	86.34±1.54&	85.87±0.57&	70.23±3.59 \\ 
			\hline
			all doctors & 78.95±2.67&	88.62±4.07&	83.23±6.2&	85.59±2.42&	74.45±5.24\\
			HSQformer-S
			&84.44±0.33&	86.5±1.27&	94.12±2.17&90.13±0.35&	86.64±0.64
			\\
			HSQformer-B
			&\textcolor{red}{85.41±0.58}&	87.62±2.07&	94.09±2.75&	90.69±0.38&	\textcolor{red}{88.32±0.59}
			\\
			HSQformer-L
			&85.38±0.6&	87.05±1.36&	\textcolor{red}{94.79±2.18}&	\textcolor{red}{90.73±0.47}&	88.23±1.08
			\\
			
		\end{tabular}
	}
	\caption{\textbf{Comparison of Radiologists and AI}. This table compares the diagnostic performance of radiologists and AI, showing that AI has surpassed the accuracy of junior radiologists and is comparable to the diagnostic level of senior radiologists.
	}
	\label{tab3}
\end{table}
\begin{figure}[h]
	\centering
	\vspace{-10pt}
	\includegraphics[width = \linewidth]{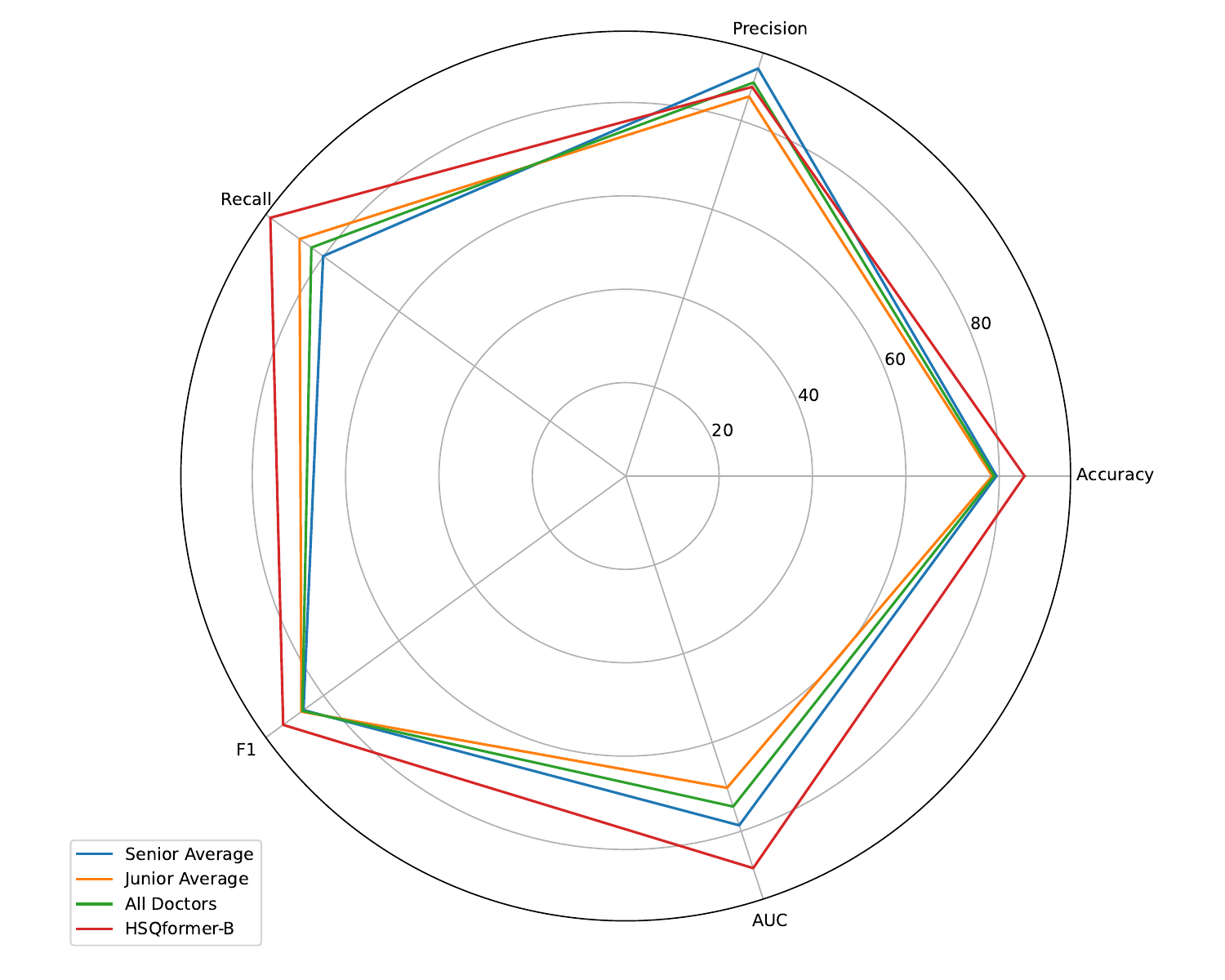}
	\caption{Visualization of Human-Machine Diagnostic Efficacy Comparison.}
	\label{fig3}
	\vspace{-10pt}
\end{figure}
In the human-machine comparison experiment, we invited six radiologists to interpret images from the high-risk patient test set, comprising three senior radiologists with over ten years of experience and three junior radiologists with 3-5 years of experience. Each radiologist was required to provide a definitive benign or malignant diagnosis for each image. To prevent potential data leakage that could bias the interpretation results, these radiologists were not involved in the data collection or exclusion process. Table \ref{tab3} and Figure \ref{fig3} detail the comparative performance of HSQformer and clinical experts in the high-risk patient screening scenario. Several intriguing observations emerged from this analysis:
\begin{itemize}
	\item Junior radiologists exhibited higher Recall (86.21\%) compared to senior radiologists (80.12\%), but lower Precision (88.62\% vs. 91.76\%). This suggests that in high-risk patient screening, junior radiologists tend to err on the side of caution by classifying uncertain cases as malignant, leading to a higher rate of false positives and thus lower Precision.
	\item HSQformer-B demonstrated a significantly higher Recall (94.09\%) than senior radiologists, with only a slight decrease in Precision (87.62\% vs. 91.76\%). Its Precision aligns closely with juniors' (87.62\% vs. 88.62\%), indicating a balance between Recall and Precision.
	\item HSQformer-B consistently outperformed junior radiologists across all metrics and matched senior radiologists. Notably, it achieved significant improvements over the overall average of all radiologists in F1 score (90.69\% vs. 85.59\%) and AUC (88.32\% vs. 74.45\%).
\end{itemize}
\begin{figure*}[t!]
	\centering
	\subfloat[{Schematic Diagram of Stage Schemes.}]{
		\includegraphics[width = 0.55\textwidth]{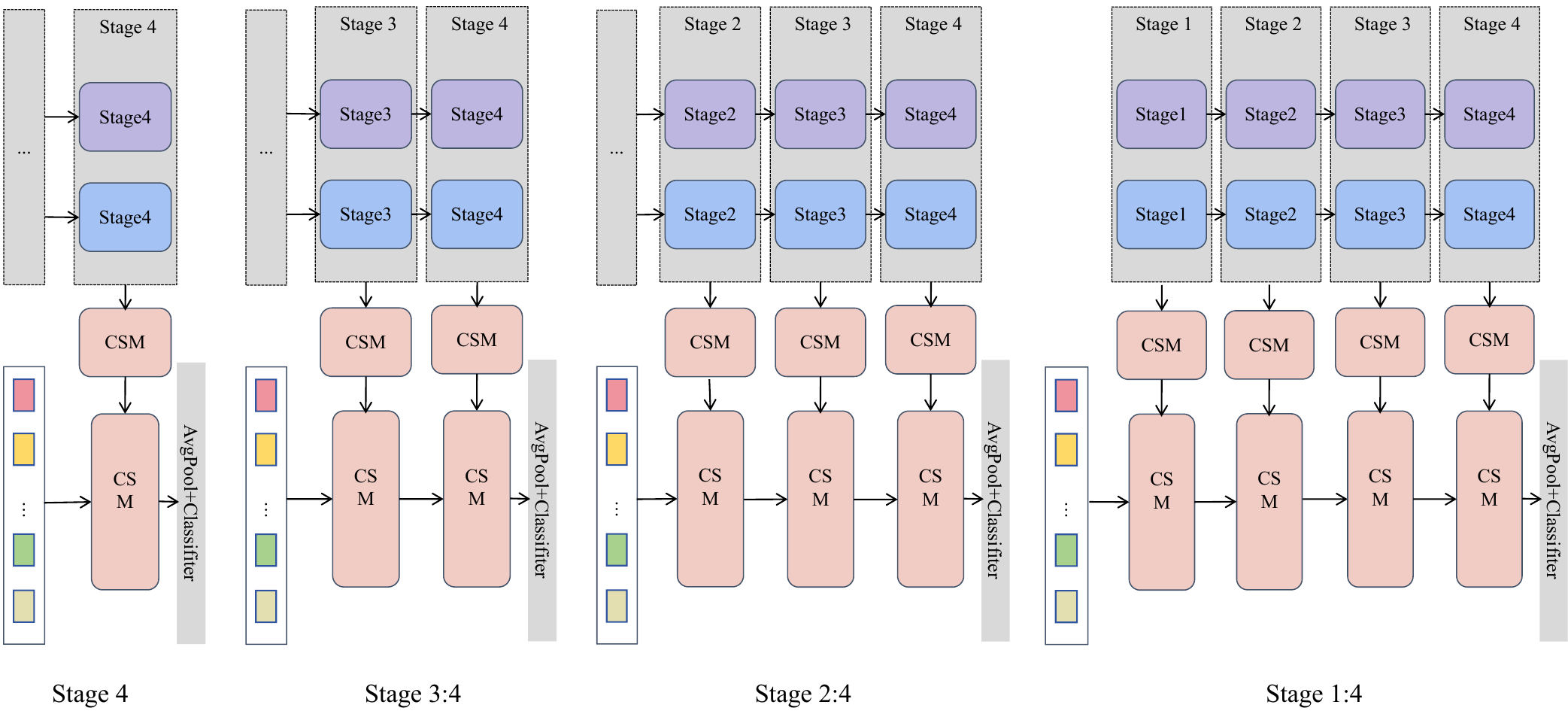}
	}
	\subfloat[{Trends in the Evolution of Stage Schemes}]{
		\includegraphics[width = 0.45\textwidth]{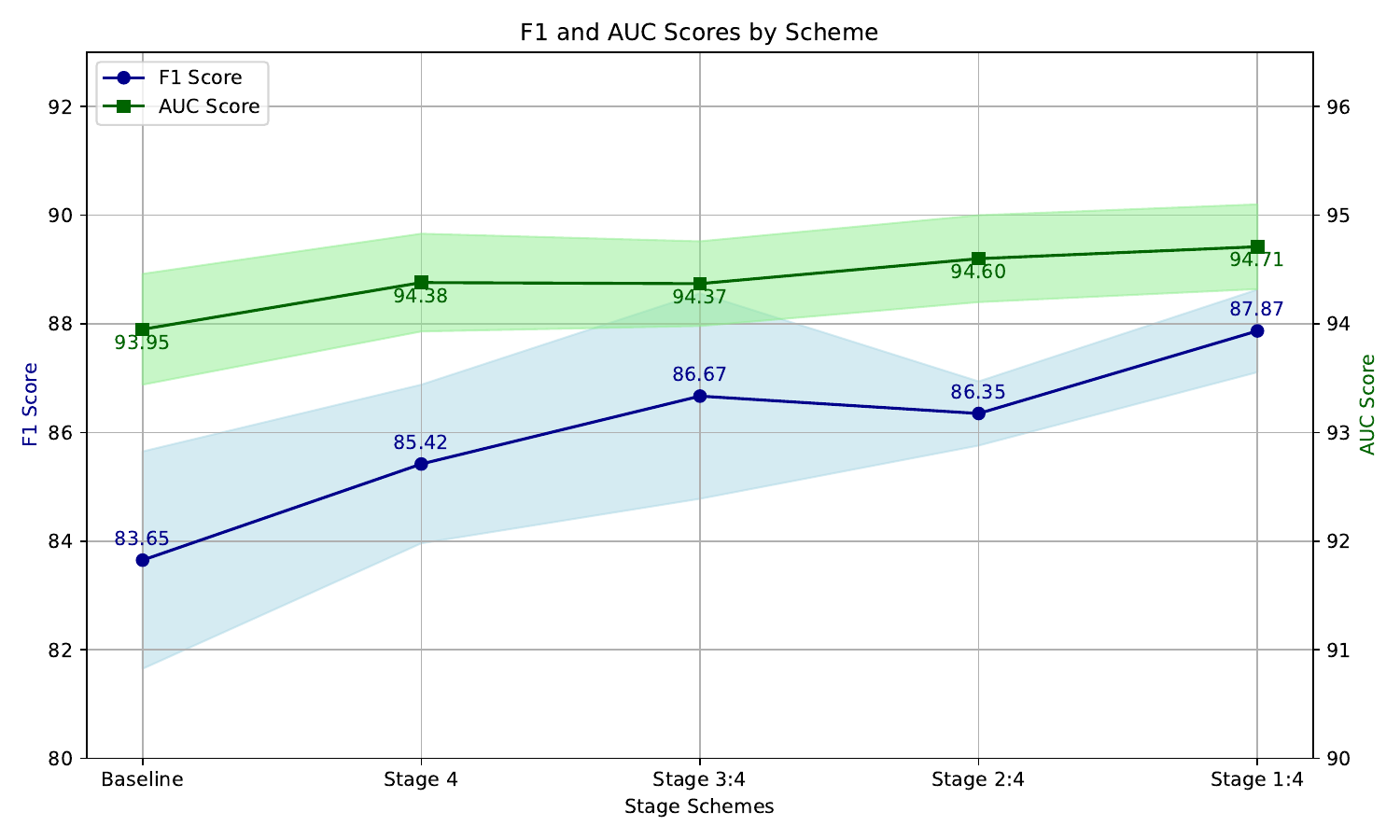}
	}
	\caption{Assessing the Impact of Stage Schemes on Model Performance. }
	\label{fig4}
	\vspace{-10pt}
	
\end{figure*}
These findings underscore HSQformer's potential as a clinical decision-support tool, especially in critical diagnostic scenarios. Its performance exceeds that of juniors and matches seniors, indicating its role in improving diagnostic accuracy and consistency. The model's strong Recall and AUC metrics highlight its effectiveness in minimizing missed diagnoses and enhancing patient outcomes.
\subsection{Ablations}
\subsubsection{Macro Ablations}
\begin{figure}[t]
	\centering
	\vspace{-10pt}
	\includegraphics[width = \linewidth]{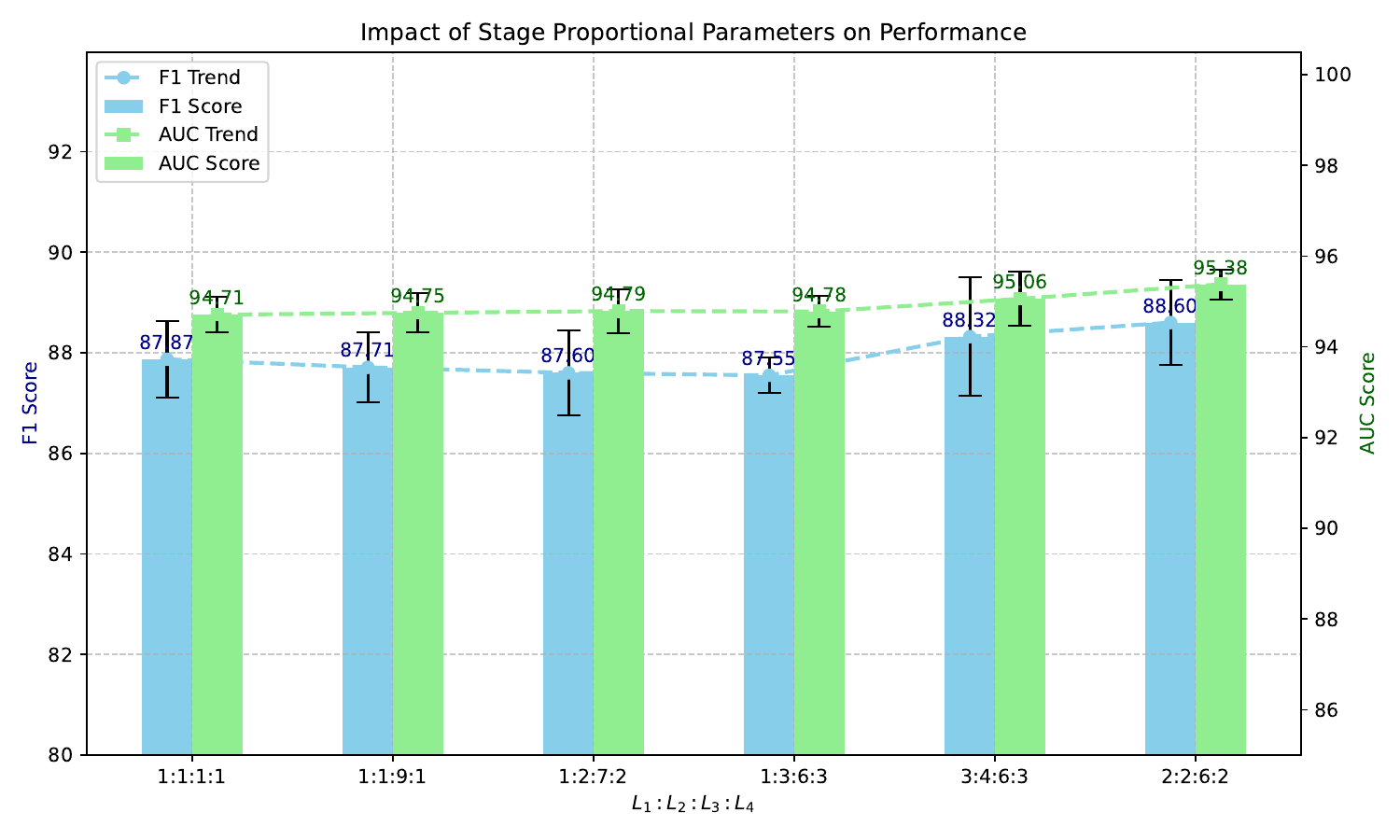}
	\caption{Influence of proportional parameters on model performance at different stages.
	}
	\label{fig5}
	\vspace{-10pt}
\end{figure}

\paragraph{Stage Schemes.} In this section, we verify the macro design of HSQformer. As illustrated in Figure \ref{fig4}, we validate the effectiveness of the hierarchical design by incrementally incorporating CSM. Unless otherwise specified, all ablation study results are obtained on the multi-center test set, as this approach better validates the generalizability of the model. We observe a progressive increase in both F1 and AUC scores. Merely adding a CSM in the final stage results in a 1.77\% improvement in F1 and a 0.43\% enhancement in AUC compared to the baseline. The full-stage approach yields a 4.22\% increase in F1 and a 0.76\% improvement in AUC over the baseline.
\paragraph{Stage ratios.} To further investigate the influence of stage ratios on model performance, we used SOTA models\cite{swin,hiera,resnet,mpvit,davit} as references and conducted extensive assessments under various proportional settings, as shown in Figure \ref{fig5}. The results indicate that the HSQformer model exhibits minimal performance variation across different stage ratios, with the AUC showing particularly slight changes. This suggests a certain level of robustness in the model's performance regarding stage ratios. However, based on the experimental data, we identified the most effective ratio configuration as 2:2:6:2, which optimizes the model's performance while balancing computational efficiency and accuracy.

\paragraph{Query number and dimension.} Within the model architecture, Query number and dimension are key parameters determining the model's efficiency and efficacy. The Query number indicates how many tokens the model can process in one pass, while the Query dimension determines each query's capacity to capture information.

\begin{figure}[t]
	\centering
	\vspace{-10pt}
	\includegraphics[width = \linewidth]{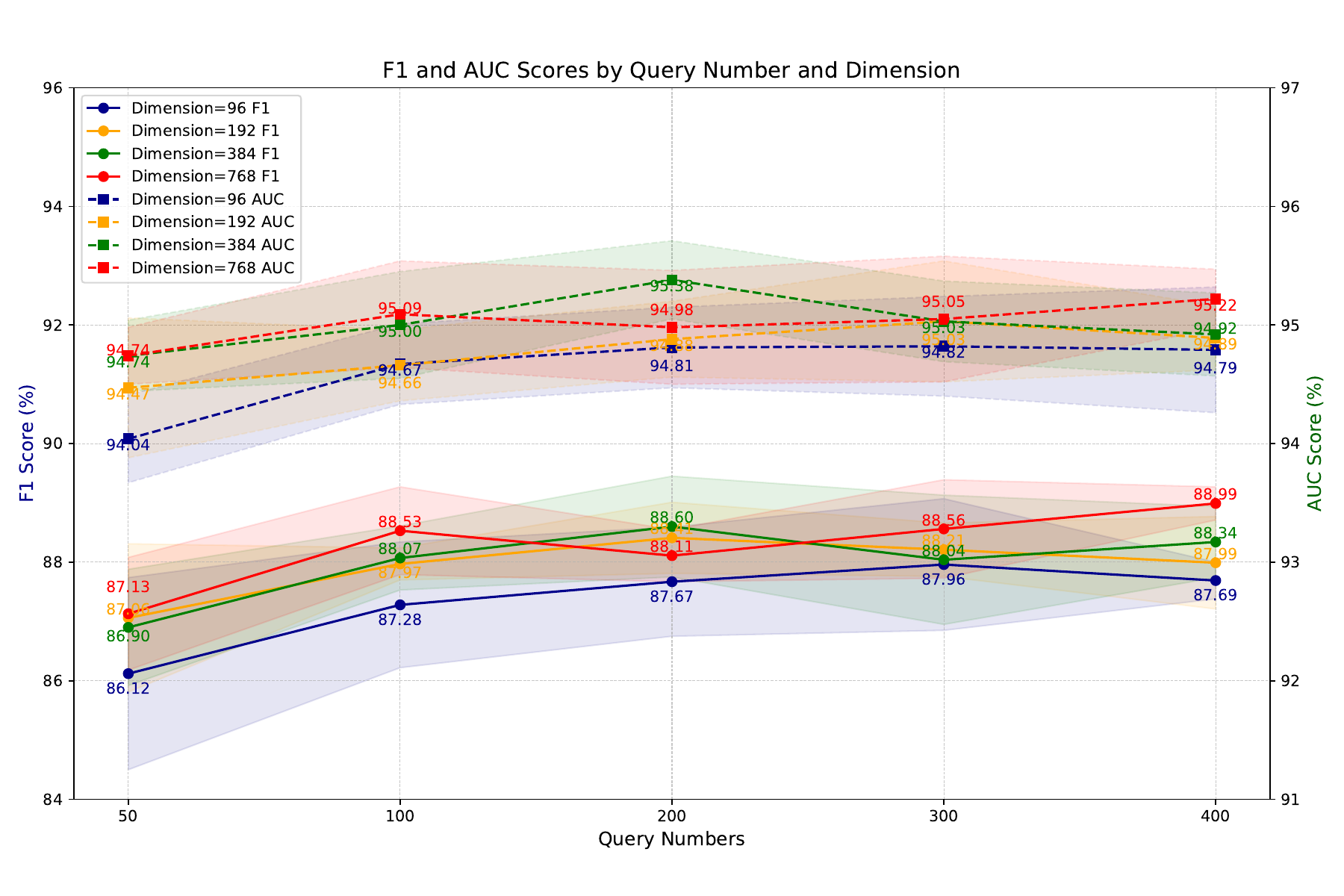}
	\caption{Systematic Analysis of Query Numbers and Dimensions on Model Performance}
	\label{fig6}
	\vspace{-10pt}
\end{figure}

Figure \ref{fig6} shows that increasing the Query dimension at a constant Query number significantly boosts the model's F1 and AUC scores. For example, with 50 queries, the F1 score rises from about 86.12\% to 87.17\%, and the AUC increases from approximately 94.04\% to 94.74\% when the Query dimension grows from 96 to 768. This means a higher Query dimension helps the model capture long - range dependencies and subtle features better, improving classification. Conversely, moderately increasing the Query number at a fixed dimension also enhances performance. At a Query dimension of 768, the F1 score goes up from about 87.13\% to 88.53\%, and the AUC increases from around 94.74\% to 95.09\% as the Query number rises from 50 to 100. However, further increasing the Query number to 300 or 400 yields little improvement. When both parameters increase concurrently, model performance can rise significantly. For instance, raising the Query number and dimension from 50 and 96 to 400 and 768 respectively, the F1 score jumps from about 86.12\% to 88.99\%, and the AUC increases from approximately 94.04\% to 95.22\%. Yet, beyond a certain point, the enhancement plateaus, indicating an optimal balance. It's crucial to note the trade - off between these parameters. Increasing either raises computational costs and model complexity. Thus, finding the optimal equilibrium is vital for achieving superior performance while maintaining computational efficiency.
\subsubsection{Micro Ablations}
\begin{figure*}[t!]
	\centering
	\subfloat[{Serial vs. Parallel Attention Mechanisms}\label{fig7(a)}]{
		\includegraphics[width = 0.6\textwidth]{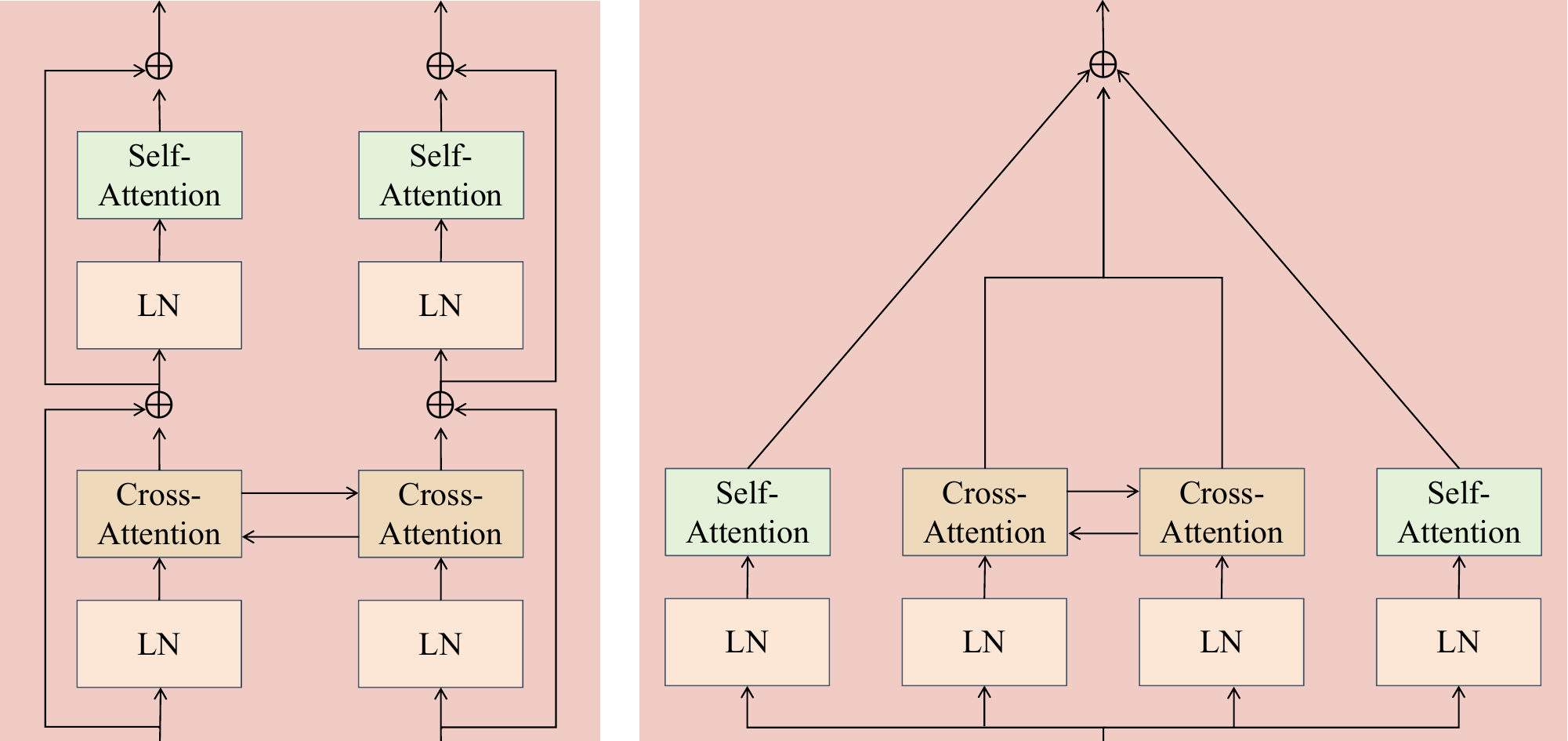}
	}
	\subfloat[{Performance Comparison}\label{fig7(b)}]{
		\includegraphics[width = 0.4\textwidth]{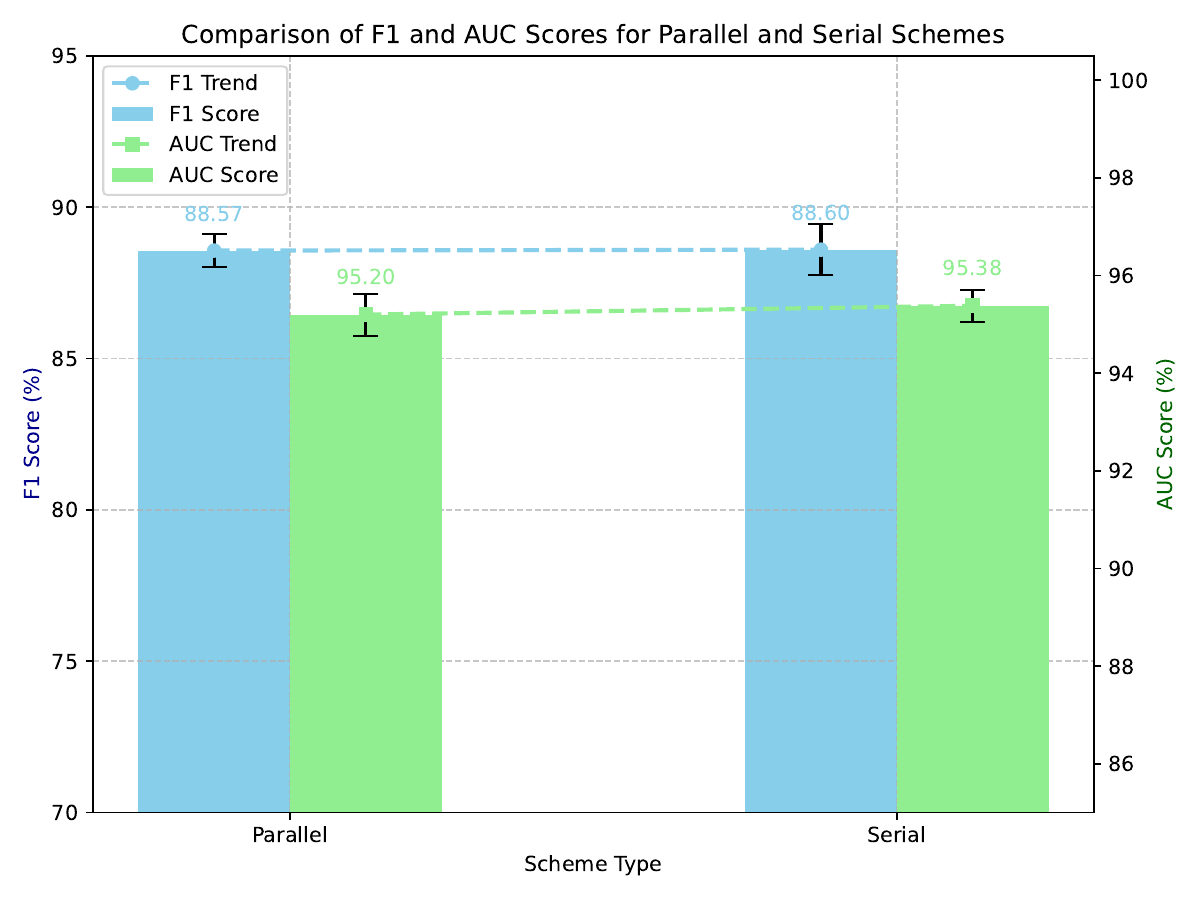}
	}
	\caption{Serial vs. Parallel Attention Mechanisms: Structural Diagram and Performance Comparison}
	\label{fig7}
	\vspace{-10pt}
\end{figure*}

\paragraph{Parallel and Serial Schemes.}
This section delves into the micro - level design efficacy of the CSM mechanism. As shown in Figure \ref{fig7}, we analyze the sequence of self - attention and cross - attention. Results reveal minimal performance difference between serial and parallel approaches. However, we choose the serial method. This is because, in this sequence, query tokens first capture relevant features through cross - attention, then self - attention reinforces these features. This sequential approach aligns with the Transformer architecture's design philosophy, ensuring our design's rationality.
\paragraph{MOE.} In the study of Mixture of Experts (MoE) models, the number of experts and the Top-K selection strategy are crucial for model performance. In sparse MoE models aiming for computational efficiency, existing research mostly uses Top-1 or Top-2 strategies. This is because selecting more experts may reduce sparsity and increase computational overhead. In contrast, dense configurations, where all experts are activated regardless of input, focus on leveraging all experts' collective knowledge for potentially better performance. 

\begin{figure}[t]
	\centering
	\vspace{-10pt}
	\includegraphics[width = \linewidth]{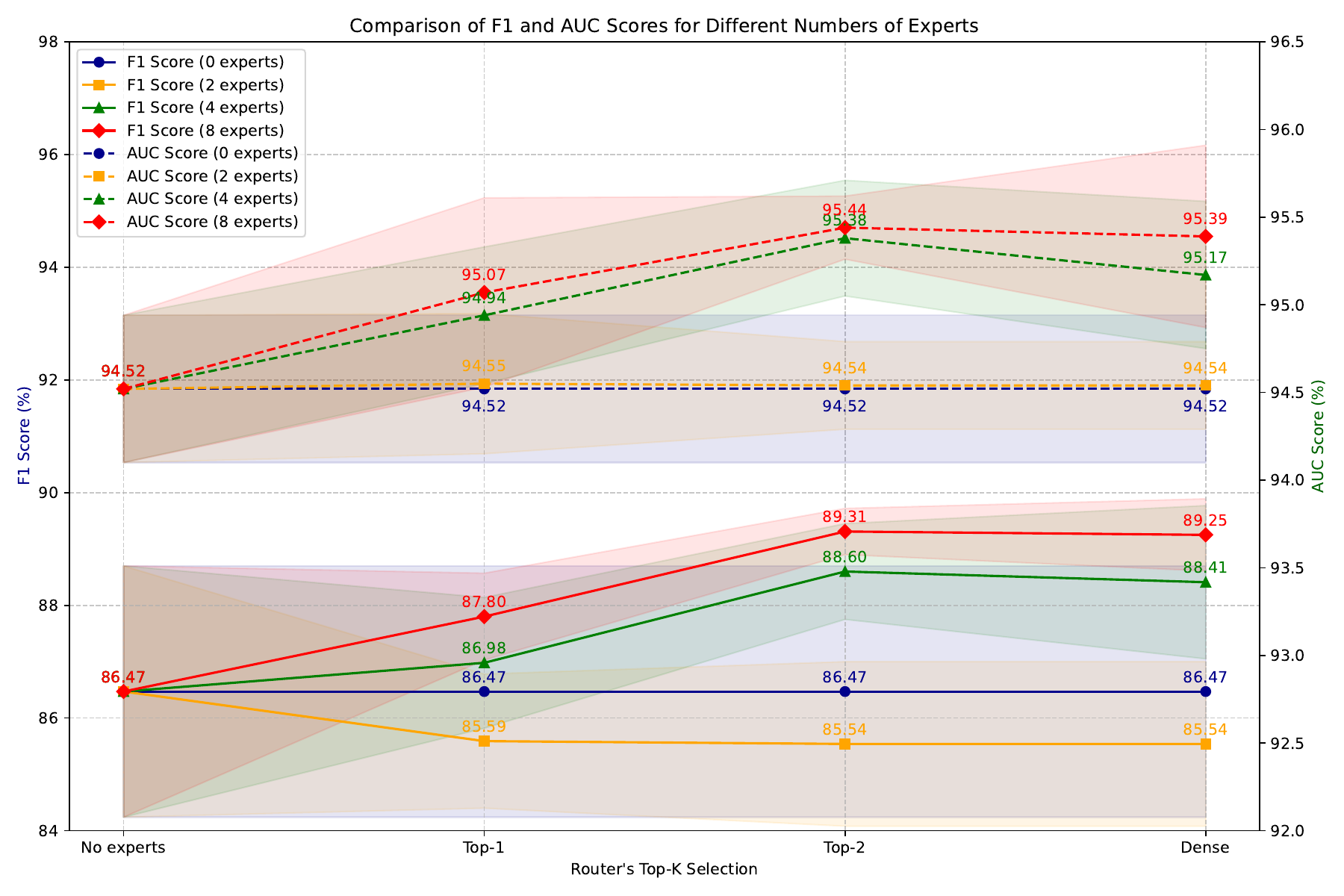}
	\caption{Trend Curves of Number of Experts and Top-K Selection Strategy
	}
	\label{fig8}
	\vspace{-10pt}
\end{figure}

Figure \ref{fig8} examines the impacts of varying expert counts and routing strategies on model performance. When the number of experts is limited, MoE shows no performance improvement over traditional MLP (which can be seen as having zero experts). Specifically, with two experts and a Top-1 strategy, MoE achieves comparable AUC scores to MLP but shows no significant F1 score improvement, with a 0.88\% decrease compared to MLP. There are also no substantial differences between Top-1 and Top-2 strategies in this case, indicating that a limited number of experts does not enhance model performance. These results suggest that when the number of experts is small ($\le 2$), MoE offers no performance benefits over conventional architectures like MLP. This highlights the need to increase the number of experts or explore alternative strategies for superior MoE performance. As the number of MoE experts increases to four or eight, significant performance improvements over MLPs are observed with a Top-1 strategy. Using a Top-2 strategy further enhances these benefits compared to Top-1. Interestingly, activating all experts in a dense MoE configuration does not lead to additional performance gains and may even cause a slight decline, potentially due to increased model complexity and overfitting. Notably, under the Top-2 strategy, MoE with four and eight experts shows similar performance, indicating that four experts might provide an optimal balance between performance and efficiency. This insight is crucial for designing efficient and high-performing MoE architectures.
\section{Discussion and Limitations}
Our research centers on evaluating AI's efficacy in HCC screening and comparing its diagnostic performance with that of clinical radiologists. The results reveal that AI outperforms junior radiologists and matches or even surpasses senior ones under specific conditions. This is consistent with recent progress in medical imaging analysis, where AI has significantly enhanced diagnostic speed and accuracy. The designed AI model shows robust generalization across three distinct scenarios, highlighting its adaptability to variations in HCC imaging caused by individual differences, tumor size, location, and growth rate. This consistency is especially vital for early-stage HCC screening, where timely diagnosis is crucial for improving patient survival rates. However, when considering AI's potential applications in HCC diagnostics, it is essential to recognize the challenges in clinical practice. Firstly, training and validating AI models require large amounts of high-quality data, which are time-consuming and costly to collect and annotate. Secondly, despite some visual insights into model predictions, as provided in the appendix, AI models' interpretability remains a challenge, particularly in explaining the decision-making process to patients and radiologists.

While our study shows promising results, several limitations must be addressed. The HSQformer, tailored for ultrasonic HCC screening, includes hyperparameters such as query number and dimension, and the number of experts and Top-K selection strategy within the MoE framework. The applicability of these parameters to other domains needs further investigation. Additionally, our study focuses on the model's efficacy without considering its parameter count. Although MoE improves performance, it also increases model size; however, only a subset of experts is activated during inference. Future work will explore using more compact models without compromising performance to enhance accessibility and broader application.
\section{Conclusion}
This study introduces an innovative Hierarchical Sparse Query Transformer model, designed to enhance the diagnostic accuracy of HCC ultrasound screening. By integrating the strengths of CNNs and ViTs, the HSQformer achieves superior performance across various clinical scenarios, including single-center, multi-center, and high-risk patient testing, significantly outperforming existing state-of-the-art models such as ConvNext and SwinTransformer. Furthermore, its diagnostic accuracy is comparable to that of senior radiologists and even surpasses junior radiologists in certain aspects. The HSQformer's success underscores the immense potential of artificial intelligence in medical image analysis, particularly in improving diagnostic efficiency and accuracy. Its modular and extensible design philosophy ensures broad applicability and flexibility in real-world clinical settings. Additionally, the open-source code facilitates future research and development, contributing to the advancement of AI in the field of HCC screening.

\section*{Acknowledgements}
This work was supported in part by the National Natural Science Foundation
of China under Grants 12326609, 82272076, 82371983 and 82171960.
{\small
	\bibliographystyle{IEEEtran}
	\bibliography{reference}
}

 \newpage
\section*{Appendix}
\setcounter{section}{0} 
\renewcommand{\thesection}{\Alph{section}} 
\section{Architecture Details} 
The HSQformer models (HSQformer-S, HSQformer-B, and HSQformer-L) are designed to scale in complexity and computational demand. The Small model offers a basic configuration, while the Base model enhances feature extraction. The Large model provides the most advanced capabilities, suitable for complex tasks. Table~\ref{tab4} summarizes their configurations.
\begin{table}[H]
	\centering
	\tablestyle{1.8pt}{1.05}
	\resizebox{1.04\linewidth}{!}{
		\begin{tabular}{l|c|c|c}
		Model	& HSQformer-S & HSQformer-B & HSQformer-L \\
			\hline
		Query number & 200 & 200 & 400 \\
			\hline
		Query dimension& 384 & 384 & 768 \\
			\hline
		Stage ratio & 1:1:1:1 & 2:2:6:2 & 2:2:6:2 \\
			\hline
		MOE type & Sparse tokens & Sparse tokens & Sparse tokens \\
			\hline
		Number of experts & 4 & 4 & 8 \\
			\hline
		Top-K Strategy	 & Top-1 & Top-2 & Top-2 \\
			\hline
		Input shape &\specialcellleft{Stage 1:$3136\times 384 $ \\ 
		 	Stage 2:$784\times 384 $ \\ 
		 	Stage 3:$196\times 384 $ \\ 
		 	Stage 4:$48\times 384 $ \\}
		  & \specialcellleft{Stage 1:$3136\times 384 $ \\ 
		  	Stage 2:$784\times 384 $ \\ 
		  	Stage 3:$196\times 384 $ \\ 
		  	Stage 4:$48\times 384 $ \\} &\specialcellleft{Stage 1:$3136\times 768 $ \\ 
		  	Stage 2:$784\times 768 $ \\ 
		  	Stage 3:$196\times 768 $ \\ 
		  	Stage 4:$48\times 768 $ \\} \\
	  		\hline
		Output shape &$200\times 384 $ & $200\times 384 $&$400\times 768 $ \\
		\end{tabular}
	}
	\caption{Configuration of Small, Base, and Large Models}
	\label{tab4}
\end{table}
\section{Training Settings} 
\begin{table}[h]
	\tablestyle{5.0pt}{1.02}
	\footnotesize
	\begin{tabular}{l|c}
		pre-training config\cite{ConvNext,swin} & ImageNet-1K 224$^2$\\
		\shline
		fine-tuning config &HSQformer-S/B/L \\
		\hline
		weight init & trunc. normal (0.2) \\
		optimizer & AdamW\\
		base learning rate & 1e-4\\
		weight decay & 1e-4 \\
		optimizer momentum & $\beta_1, \beta_2{=}0.9, 0.999$ \\
		batch size & 50(S-B)/20(L)  \\
		fine-tuning epochs & 20 \\
		learning rate schedule & cosine decay  \\
		warmup epochs & 1  \\
		warmup schedule & cosine \\
		layer-wise lr decay  & None \\
		auto augment  & IMAGENET  \\
		random resized crop  & 224/(0.7,1.0)  \\
		random horizontal flip  & 0.5  \\
		mixup  & 0.8 \\
		cutmix  & 1.0  \\
		random erasing  & 0.25  \\
		label smoothing  & 0.1  \\
		stochastic depth  & 0.2  \\
		dropout	& 0.5  \\
		layer scale  & 1e-6 \\
		head init scale & None  \\
		gradient clip & None\\
		exp. mov. avg. (EMA) & None(S-B)/0.9999(L)\\
	\end{tabular}
	\caption{Detailed overview of the training configurations.}
	\label{tab:train_detail}
\end{table}
Table~\ref{tab:train_detail} provides a detailed overview of the configurations used in both the pre-training and fine-tuning stages. The pre-training phase adheres to the protocol described in~\cite{cswin}, with the primary objective of enabling the model to learn a robust set of features from a large-scale dataset. The fine-tuning stage builds upon the pre-trained model by adjusting various parameters to further enhance its performance, ensuring it is better suited to specific task requirements. This optimization strategy, transitioning from pre-training to fine-tuning, helps improve the model's generalization and task adaptability.
\section{Visualization}
\begin{figure*}[h]
 	\centering
 	\includegraphics[width=0.9\linewidth]{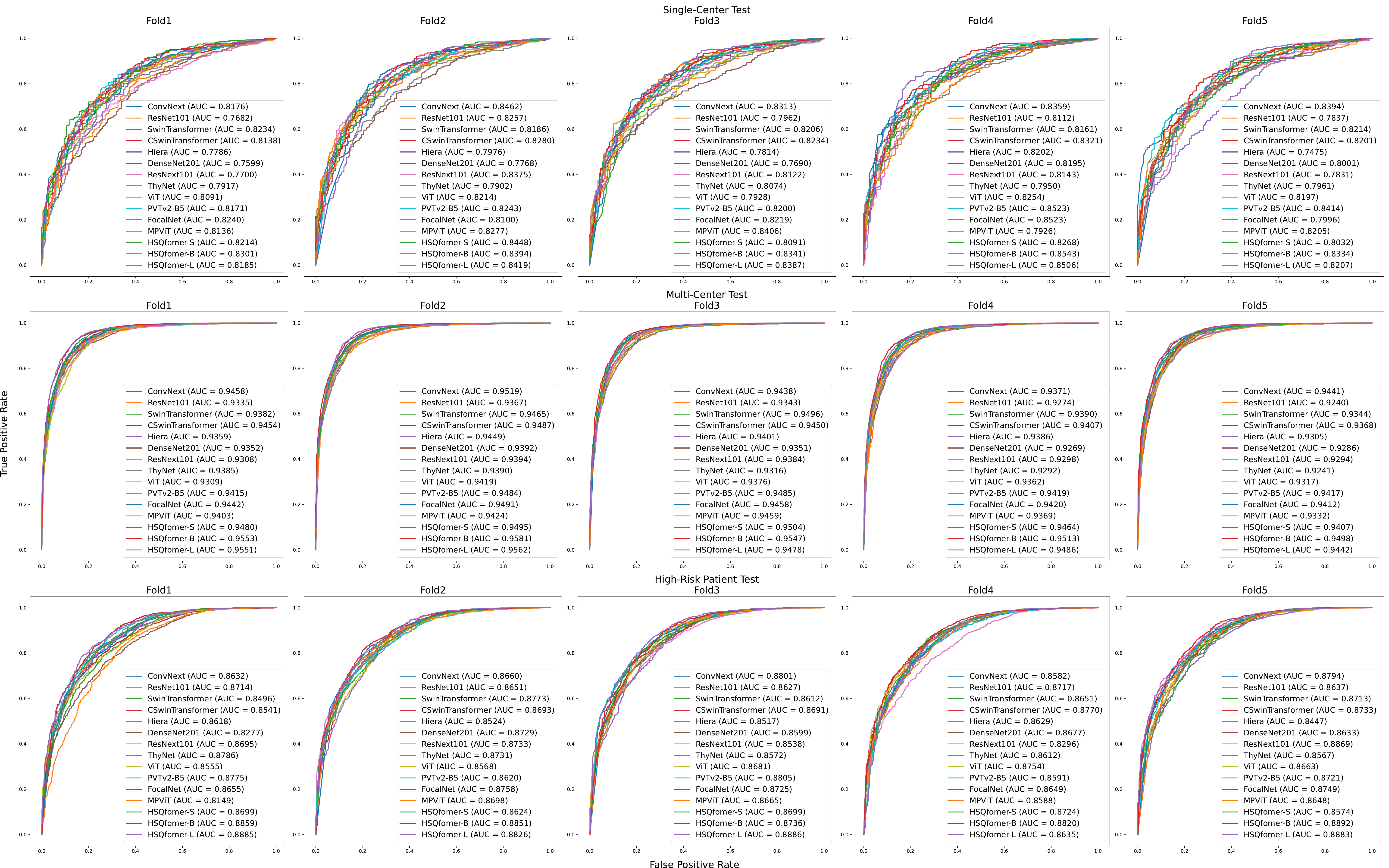}
 	\caption{Systematic Comparative Analysis of SOTA Models Using 5-Fold Cross-Validation on Three Test Datasets
 	}
 	\label{fig9}
 \end{figure*}

\begin{figure*}[h]
 	\centering
 	\includegraphics[width=0.9 \linewidth]{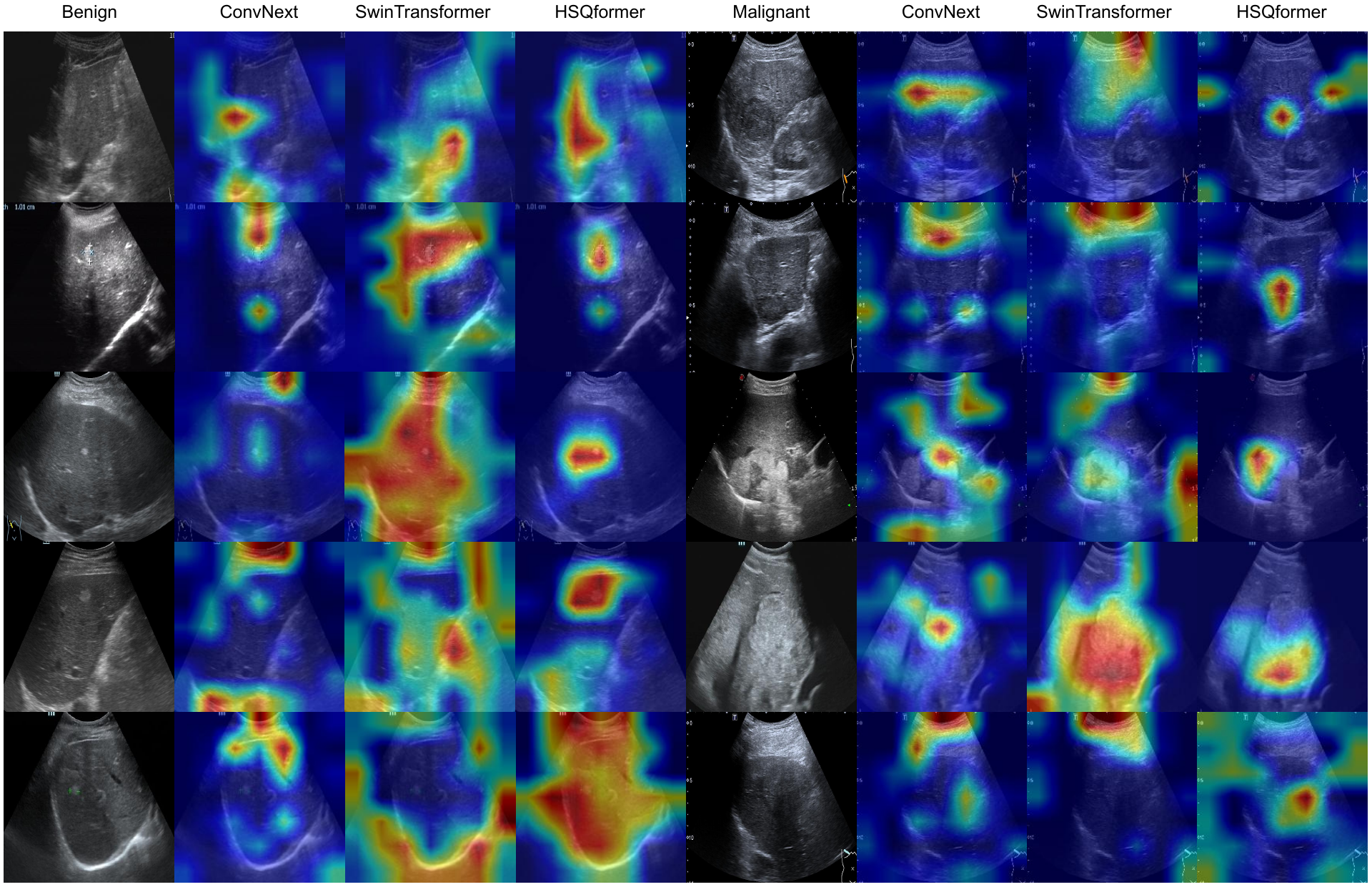}
 	\caption{Heatmap Analysis of Attention Maps for Ultrasound HCC Diagnosis. 
 	}
 	\label{fig10}
 \end{figure*}

In Figure \ref{fig9}, we conducted a systematic comparative analysis of the performance of state-of-the-art models on three different test datasets using a 5-fold cross-validation method. The figure provides an intuitive visualization of the models' performance across various testing scenarios, thereby aiding in a deeper understanding of the strengths and limitations of each model. Figure \ref{fig10}, on the other hand, offers a hotspot analysis in the form of a heatmap. This heatmap visualization highlights the regions of interest in the ultrasound images where the models focus their attention for making predictions. The intensity of the colors in the heatmap corresponds to the importance of the respective regions, with brighter colors indicating higher significance. This interpretability analysis helps us understand the underlying patterns and features that the models rely on for diagnosing HCC, providing insights into their decision-making process.

\end{document}